\documentclass[11pt]{article}
\usepackage[hyperref]{acl}
\usepackage{times}
\usepackage{latexsym}
\usepackage{amsmath}
\usepackage{amssymb}
\usepackage{algorithm}
\usepackage{algorithmic}
\usepackage{graphicx}
\usepackage{booktabs}
\usepackage{multirow}
\usepackage{xcolor}
\usepackage{bbding}

\usepackage[table]{xcolor}

\definecolor{Highlight}{rgb}{0.92,0.94,1}
\definecolor{cream}{RGB}{255, 251, 234}
\definecolor{goldenstar}{RGB}{255, 204, 0}
\usepackage{tikz}

\usepackage[most]{tcolorbox}
\tcbuselibrary{skins, breakable}
\usepackage{xcolor}
\definecolor{goldenstar}{RGB}{255, 204, 0}
\usepackage{fontawesome}

\usepackage{tcolorbox}
\tcbuselibrary{skins, breakable}  

\definecolor{cream}{RGB}{255, 253, 245}
\definecolor{goldenstar}{RGB}{255, 193, 7}

\tcbset{
  takeawaystyle/.style={
    colback=Highlight,
    colframe=black!30,
    boxrule=0.4pt,
    arc=3pt,
    left=6pt,
    right=6pt,
    top=6pt,
    bottom=6pt,
    fonttitle=\bfseries,
    coltitle=black,
    enhanced,
    drop shadow
  },
  promptstyle/.style={  
    colback=blue!5!white,
    colframe=blue!75!black,
    boxrule=0.5pt,
    arc=2pt,
    left=8pt,
    right=8pt,
    top=8pt,
    bottom=8pt,
    fonttitle=\bfseries\sffamily,
    coltitle=white,
    enhanced,
    attach boxed title to top left={yshift=-2mm, xshift=5mm},
    boxed title style={
      colback=blue!75!black,
      arc=2pt,
    },
  }
}

\usepackage{xspace}
\newcommand{\mname}{{\texttt{InstructDiff}}\xspace}
\newcommand{\equalcontrib}{\textsuperscript{*}}
\newcommand{\corresponding}{\textsuperscript{\Envelope}}

\title{\mname: Domain-Adaptive Data Selection via Differential Entropy for Efficient LLM Fine-Tuning}

\author{
  Junyou Su$^{1\equalcontrib}$ \; 
  He Zhu$^{1\equalcontrib\corresponding}$ \; 
  Xiao Luo$^{2}$ \; 
  Liyu Zhang$^{1}$ \; 
  Hong-Yu Zhou$^{3}$ \\ 
  \textbf{Yun Chen}$^{4}$ \; 
  \textbf{Peng Li}$^{3}$ \; 
  \textbf{Yang Liu}$^{3}$ \; 
  \textbf{Guanhua Chen}$^{5\corresponding}$ \\
  $^{1}$Peking University \; $^{2}$University of Wisconsin-Madison \\
  $^{3}$Tsinghua University \; $^{4}$SUFE \; $^{5}$SUSTech \\
  \texttt{\{jysu25, zhuhe\}@stu.pku.edu.cn}, \texttt{chengh3@sustech.edu.cn}
}

\begin{document}
\maketitle 


\insert\footins{
  \noindent\footnotesize 
  $^*$Equal contribution. \quad 
  \Envelope Corresponding authors. \\
  GitHub: \href{https://github.com/zhuchichi56/Instruct-diff}{https://github.com/zhuchichi56/Instruct-diff}
}
\def\thefootnote{\arabic{footnote}}

\begin{abstract}
Supervised fine-tuning (SFT) is fundamental to adapting large language models, yet training on complete datasets incurs prohibitive costs with diminishing returns. Existing data selection methods suffer from severe domain specificity: techniques optimized for general instruction-following fail on reasoning tasks, and vice versa. We observe that measuring entropy differences between base models and minimally instruction-tuned calibrated models reveals a pattern—samples with the \textit{lowest differential entropy} consistently yield optimal performance across domains, yet this principle manifests domain-adaptively: reasoning tasks favor entropy \textit{increase} (cognitive expansion), while general tasks favor entropy \textit{decrease} (cognitive compression). We introduce \mname, a unified framework that operationalizes differential entropy as a domain-adaptive selection criterion through warmup calibration, bi-directional NLL filtering, and entropy-based ranking. Extensive experiments show that \textsc{\mname} achieves 17\% relative improvement over full data training on mathematical reasoning and 52\% for general instruction-following, outperforming prior baselines while using only 10\% of the data.
\end{abstract}

\section{Introduction}
\label{sec:introduction}
Recent advances in large language models have demonstrated remarkable capabilities, achieving unprecedented performance on complex reasoning and language understanding tasks~\cite{openai2025o3,guo2025deepseek,anthropic2025claude}. Post-training has emerged as the predominant paradigm for unlocking these capabilities~\cite{ouyang2022training,dubey2024llama}, with supervised fine-tuning (SFT) serving as the cornerstone stage that shapes model behavior through instruction-response pairs. As post-training datasets continue to expand, often reaching millions of examples~\cite{xu2024magpie}, a critical challenge arises: training on entire datasets incurs prohibitive computational costs while delivering diminishing returns. Recent evidence demonstrates that \textit{data quality fundamentally dictates both training efficiency and model performance}~\cite{zhou2024lima,ye2025limo,xiao2025limi}. Carefully curated subsets can achieve comparable or superior results with dramatically fewer examples, reducing training time by orders of magnitude. This observation establishes \textit{data curation}, the systematic selection of high-quality training examples, as an essential strategy for efficient and effective model adaptation.

\begin{figure*}[t]
    \centering
    \begin{minipage}[b]{0.33\textwidth}
        \centering
        \includegraphics[width=\textwidth]{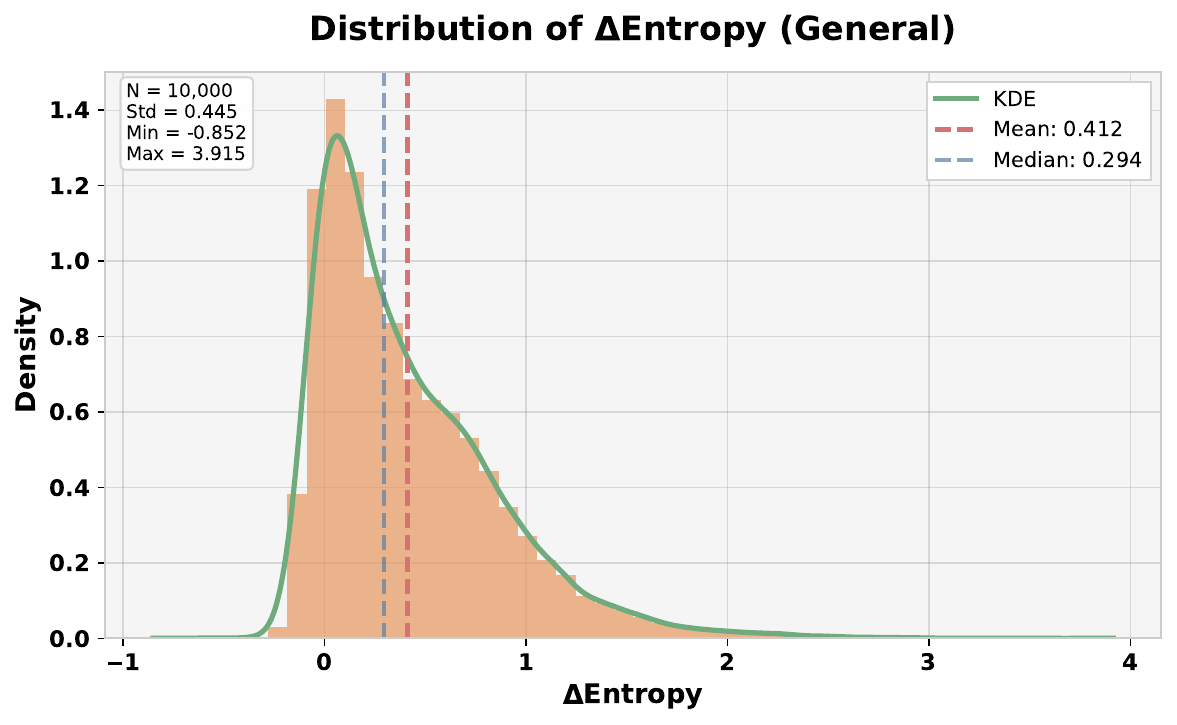}
    \end{minipage}\hfill
    \begin{minipage}[b]{0.33\textwidth}
        \centering
        \includegraphics[width=\textwidth]{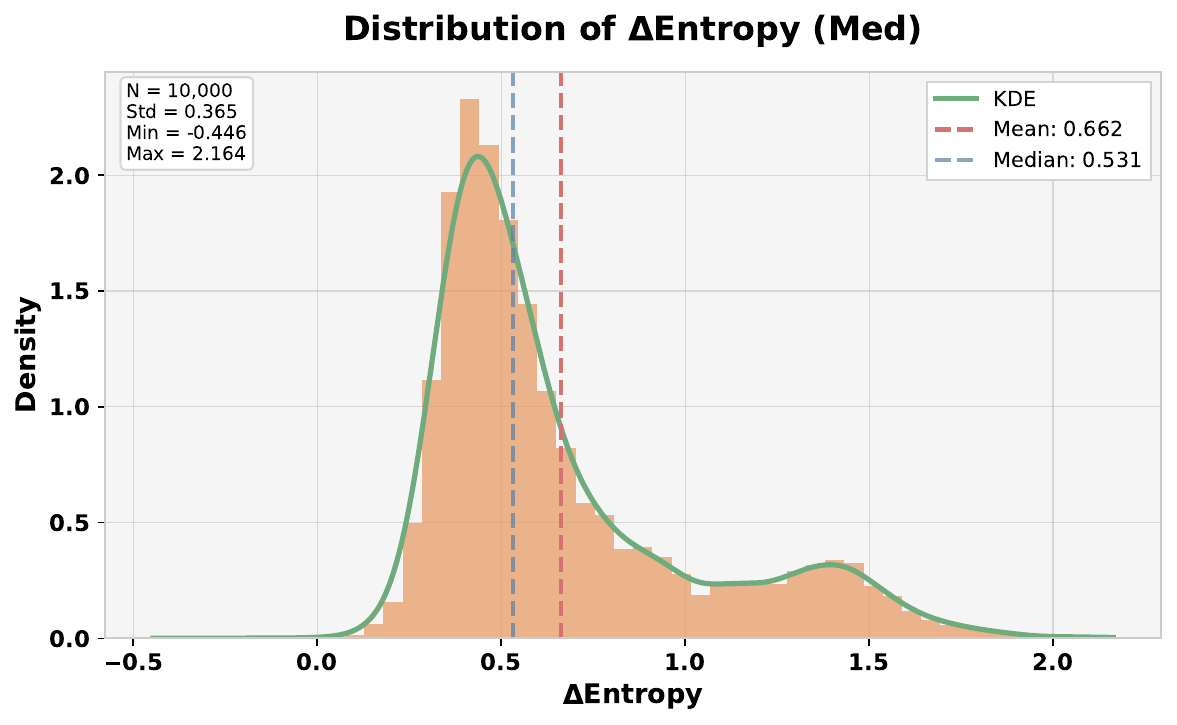}
    \end{minipage}\hfill
    \begin{minipage}[b]{0.33\textwidth}
        \centering
        \includegraphics[width=\textwidth]{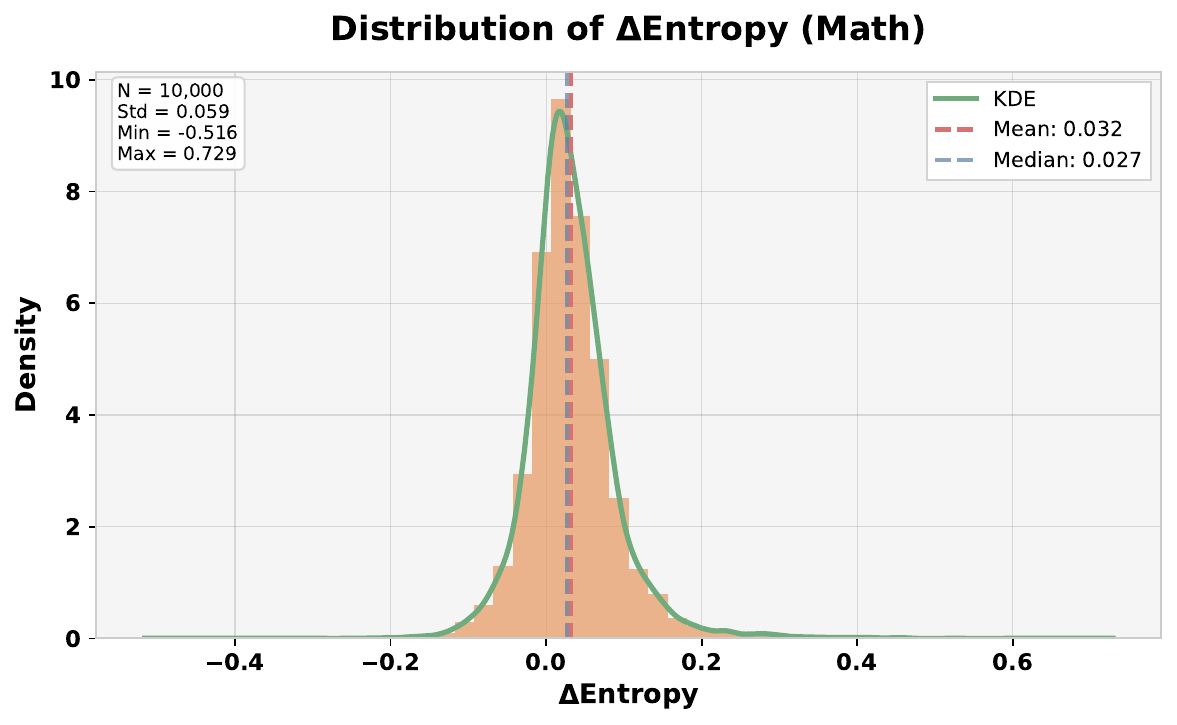}
    \end{minipage}
    \caption{\textbf{Domain-adaptive entropy dynamics reveal distinct learning patterns.} We visualize the entropy difference ($\Delta H = H_{\text{base}} - H_{\text{inst}}$) distributions across three domains. For \textbf{general instruction-following} and \textbf{medical QA}, almost all samples show $H_{\text{base}} > H_{\text{inst}}$, i.e., instruction-tuning compresses uncertainty (\textbf{entropy decrease}, cognitive compression). In contrast, for \textbf{mathematical reasoning}, about half of the samples show $H_{\text{inst}} > H_{\text{base}}$, i.e., instruction-tuning \emph{increases} entropy (cognitive expansion). This contrast supports our unified but domain-adaptive selection principle based on differential entropy.}
    \label{fig:domain_entropy_dynamics_small}
\end{figure*}

Current data selection methods exhibit severe domain specificity, partitioning into siloed techniques that struggle to generalize across task structures. Existing approaches broadly fall into two categories: methods for \textbf{general tasks} that prioritize conversational quality through heuristics such as influence functions, quality metrics, or uncertainty~\cite{xia2024less,li2024quantity,liu2024selectit}, and methods for \textbf{reasoning tasks} that prioritize verifiable correctness through difficulty-based filtering, Best-of-N sampling, or pass@k optimization~\cite{lightman2023let,liu2025acemath,lyu2025top,walder2025pass}. While these domain-specific strategies have demonstrated effectiveness within their respective domains, their applicability across varied tasks remains somewhat constrained. This raises a fundamental question: \textit{Is it possible to devise a unified, domain-adaptive method or metric for data selection that automatically adjusts to the characteristics of different tasks and domains?}

To address this challenge, we re-examine fine-tuning through an information-theoretic lens and uncover an empirical pattern: when comparing the base model with a \textit{calibrated model} (obtained by fine-tuning on a small random subset), we find that selecting data with the \textit{lowest differential entropy} consistently yields optimal results across domains (see Figure~\ref{fig:entropy_range} for systematic range analysis). However, how this \textit{Lowest Differential Entropy} principle appears depends fundamentally on the task structure (Figure~\ref{fig:domain_entropy_dynamics_small}). For \textbf{reasoning tasks}, the lowest differential entropy corresponds to \textit{negative} values, meaning that calibration \textit{increases} entropy. We term this pattern \textbf{cognitive expansion} and it reflects that effective reasoning requires preserving solution path diversity. In contrast, for \textbf{general tasks} such as instruction-following or medical tasks, the lowest differential entropy corresponds to \textit{small positive} values, where calibration \textit{decreases} entropy. This is called \textbf{cognitive compression} and fine-tuning in this case concentrates diffuse priors onto canonical behaviors. Despite these differences, both regimes follow the same principle: selecting samples with minimal absolute entropy shifts. For reasoning, these minimal shifts manifest as expansion (negative $\Delta H$), while for general tasks they appear as compression (small positive $\Delta H$).

Building on this insight, we introduce \textbf{\mname}, a model-aware data selection framework that leverages model-state difference metrics as a unified, domain-adaptive criterion for supervised fine-tuning (SFT). \textsc{\mname} is designed as a two-stage process. In the first stage, \textit{Warmup Calibration}, we fine-tune the base model on a small random subset to obtain a calibration model that serves as an instruction-tuned reference. In the second stage, \textit{Distribution-Aware Selection}, we compute, for each sample, both the negative log-likelihood difference ($\Delta$NLL) and entropy difference ($\Delta H$) between the base and calibration models. We apply bi-directional NLL filtering to remove both redundant and incomprehensible samples, then rank the remaining data by $\Delta H$ and select those with the lowest values from the learnable range. This approach allows a single criterion to naturally adapt. For reasoning tasks, the method favors samples with entropy increase (cognitive expansion), and for general instruction-following, it favors entropy decrease (cognitive compression). The entire procedure requires no explicit domain heuristics. Furthermore, we support an efficient \textit{iterative selection} strategy. After training on the initially selected data, the workflow can be repeated with the updated model to achieve further performance gains. Extensive experiments show that \textsc{\mname} achieves 17\% relative improvement over full data training on mathematical reasoning and 52\% for general instruction-following, outperforming prior baselines while using only 10\% of the data.

\section{Related Work}
\subsection{Data Selection for Supervised Fine-Tuning}
Efficient data selection is central to LLM adaptation, but most existing approaches, whether task-specific or model-agnostic, struggle to generalize across domains~\cite{wettig2024qurating}. Early studies~\cite{zhou2024lima,ye2025limo} demonstrated that carefully chosen small subsets, sometimes as few as 1,000 examples, can match or surpass full-data fine-tuning, underscoring the value of data quality over sheer quantity. Representative methods include \textbf{LLM-as-judge} scoring~\cite{chen2023alpagasus,liu2023makes}, which leverages advanced models to rate data but is costly and decoupled from the specific target model. Other approaches are \textbf{difficulty- or uncertainty-based}~\cite{li2024quantity,li2024superfiltering,zhu2025taginstruct,lu2023instag}, using single-model metrics like loss, perplexity, or instruction complexity based on tag counts; however, these often struggle to adapt across models or domains. \textbf{Gradient/influence-based} techniques~\cite{xia2024less,li2024one,li2023self} estimate training influence but require significant computation. Heuristic strategies are also common, such as ranking by length~\cite{zhu2024fanno,xia2024rethink}. To address these challenges, we propose quantifying \textit{distributional change}, specifically the negative log-likelihood and entropy differences ($\Delta$NLL, $\Delta H$) between the base model and a lightly instruction-tuned reference, as a unified and adaptive data selection criterion that is both scalable and robust across diverse SFT scenarios.

\subsection{Entropy in Language Model Training}

Entropy quantifies output uncertainty in LLMs and plays dual roles. ZIP~\citep{yin2024entropy} exploits an entropy-based compression ratio to select low-redundancy, high-diversity data, improving training efficiency. In RL, entropy regularization promotes exploration, and recent work exploits semantic- and token-level entropy to improve reasoning LLMs via adaptive entropy bonuses or high-entropy decision weighting~\citep{vanlioglu2025entropy,wang2025beyond}.In inference, entropy has been used to characterize uncertainty at the sequence level and to detect hallucinated outputs through semantic uncertainty estimation~\citep{farquhar2024detecting,zhu2025uncertainty}. Critically, prior work examines entropy within a \textit{single model state}. Our work pioneers \textbf{entropy difference} ($\Delta H$) across model states, revealing domain-specific patterns: negative $\Delta H$ (entropy increase) for reasoning versus positive $\Delta H$ (decrease) for general tasks. This shift from absolute to \textit{differential entropy dynamics} enables automatic domain adaptation.

\begin{figure*}[t]
    \centering
    \includegraphics[width=0.999\textwidth]{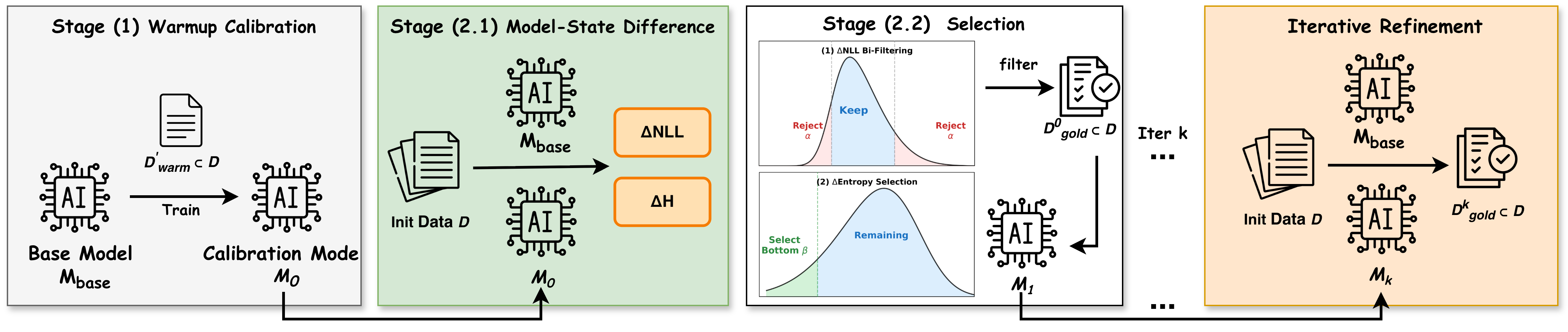}
    \caption{\textbf{The \mname{} two-stage selection pipeline.} \textbf{Stage 1: Warmup Calibration.} Randomly sample a small warmup subset to lightly instruction-tune the base model, producing a calibration model as reference. \textbf{Stage 2: Distribution-Aware Selection.} For each candidate sample, compute negative log-likelihood difference ($\Delta$NLL) and entropy difference ($\Delta H$) between the base and calibration models. Filter out samples with extreme $\Delta$NLL, then select the lowest $\Delta H$ samples from the learnable range, in a domain-adaptive way. This process can optionally be repeated with the updated model for further gains.}
    \label{fig:pipeline}
\end{figure*}

\section{Methodology}
\label{sec:methodology}

In this section, we formalize the data selection problem and describe our two-stage selection framework with iterative refinement. Figure~\ref{fig:pipeline} illustrates the complete pipeline.

\subsection{Problem Formulation and Motivation}
\label{sec:formulation}

\noindent\textbf{Formulation.} Let $\mathcal{D} = \{(x_i, y_i)\}_{i=1}^N$ denote a supervised fine-tuning dataset, where $x_i$ represents the instruction and $y_i$ the target response. Given a pre-trained base model $\pi_{\text{base}}$ and computational budget $k \ll N$, our objective is to select a subset $\mathcal{D}' \subset \mathcal{D}$ with $|\mathcal{D}'| = k$ that maximizes downstream performance while mitigating emergent misalignment~\cite{betley2025emergent}.

\noindent\textbf{Motivation.} Our empirical investigation (Figure~\ref{fig:domain_entropy_dynamics_small}) reveals that comparing $\pi_{\text{base}}$ with a minimally instruction-tuned model $\pi_{\text{inst}}$ uncovers domain-adaptive entropy patterns. For reasoning tasks, optimal samples exhibit entropy \textit{increase} (cognitive expansion); for general tasks, entropy \textit{decrease} (cognitive compression). Critically, both regimes converge on selecting samples with the \textit{lowest} differential entropy. This observation motivates operationalizing $\Delta H$ between $\pi_{\text{base}}$ and $\pi_{\text{inst}}$ as a unified, domain-adaptive selection criterion.

\subsection{Model-State Difference Metrics}
\label{sec:state-diff}

To operationalize our selection principle, we define two complementary metrics that capture distributional changes between $\pi_{\text{base}}$ and $\pi_{\text{inst}}$. For each sample $(x_i, y_i)$, the negative log-likelihood (NLL) difference quantifies learning signal strength:
\begin{equation}
    \Delta\text{NLL}_i = \mathcal{L}_{\text{inst}}(x_i, y_i) - \mathcal{L}_{\text{base}}(x_i, y_i)
\label{eq:delta_nll}
\end{equation}
where $\mathcal{L}_{\pi}(x, y) = -\frac{1}{|y|} \sum_{t=1}^{|y|} \log \pi_{\theta}(y_t \mid x, y_{<t})$ denotes length-normalized NLL. The entropy difference reveals the mode of learning:
\begin{equation}
    \Delta H_i = H_{\text{base}}(x_i, y_i) - H_{\text{inst}}(x_i, y_i)
\label{eq:delta_h}
\end{equation}
where per-token entropy is:
\begin{equation}
\resizebox{0.89\linewidth}{!}{
    $H_{\pi}(x, y) = -\frac{1}{|y|} \sum_{t=1}^{|y|} \sum_{v \in \mathcal{V}} \pi_{\theta}(v \mid x, y_{<t}) \log \pi_{\theta}(v \mid x, y_{<t})$
}
\end{equation}

Positive $\Delta H$ indicates uncertainty compression during fine-tuning, while negative $\Delta H$ suggests solution diversity expansion. Together, $\Delta\text{NLL}$ identifies the learnable range, and $\Delta H$ enables domain-adaptive selection within that range.

\subsection{Two-Stage Selection Framework}
\label{sec:two-stage}

\textsc{\mname} implements our selection principle through two stages (Figure~\ref{fig:pipeline}).
\\
\\
\noindent\textbf{Stage (1) Warmup Calibration:}
To obtain the reference distribution $\pi_{\text{inst}}$, we randomly sample a warmup subset $\mathcal{D}_{\text{warmup}}$ of size $\alpha \cdot N$ (typically $\alpha = 0.1$) and fine-tune:
\begin{equation}
\resizebox{0.89\linewidth}{!}{
    $\pi_{\text{inst}}^{(0)} = \arg\min_{\pi} \mathbb{E}_{(x,y) \sim \mathcal{D}_{\text{warmup}}} \left[ -\log \pi_{\theta}(y \mid x) \right]$
}
\end{equation}
Recent work demonstrates that small subsets suffice to activate latent instruction-following capabilities~\cite{zhou2024lima}, making this lightweight calibration both feasible and effective.
\\
\\
\noindent\textbf{Stage (2) Distribution-Aware Selection:} We compute $\Delta\text{NLL}_i$ and $\Delta H_i$ for all samples using Equations~\eqref{eq:delta_nll} and~\eqref{eq:delta_h}, then apply bi-directional NLL filtering followed by entropy-based ranking.

\textbf{Bi-Directional NLL Filtering.} We exclude pathological extremes through symmetric rejection: the bottom $\gamma$ percentile (typically $\gamma = 0.1$) contains near-duplicates providing negligible gradient signal, while the top $\gamma$ percentile contains incomprehensible patterns that may induce catastrophic forgetting~\cite{kirkpatrick2017overcoming}. This retains the learnable middle range:
\begin{equation}
\resizebox{0.89\linewidth}{!}{
$\mathcal{D}_{\text{filtered}} = \left\{ (x_i, y_i) \in \mathcal{D} : q_{\gamma} \leq \Delta\text{NLL}_i \leq q_{1-\gamma} \right\}$
}
\end{equation}

\textbf{Entropy-Based Selection:} From $\mathcal{D}_{\text{filtered}}$, we select $\beta \cdot N$ samples (typically $\beta = 0.1$) by ranking according to $\Delta H$. Empirical validation (Section~\ref{subsec:entropy-guided}) confirms that selecting samples with the \textit{lowest} $\Delta H$ consistently yields superior performance. This unified criterion automatically adapts: negative $\Delta H$ (cognitive expansion) for reasoning tasks, and low positive $\Delta H$ (cognitive compression) for general tasks, without requiring manual domain specification.

\subsection{Iterative Refinement}
\label{sec:iterative-refinement}

After fine-tuning on the initially selected subset $\mathcal{D}'^{(1)}$, we obtain an improved calibration model:
\begin{equation}
\resizebox{0.88\linewidth}{!}{
    $\pi_{\text{inst}}^{(k)} = \arg\min_{\pi} \mathbb{E}_{(x,y) \sim \mathcal{D}'^{(k)}} \left[ -\log \pi_{\theta}(y \mid x) \right]$
}
\end{equation}
This refined model serves as the measuring instrument for iteration $k+1$, enabling progressive identification of samples at the model's evolving learnable frontier. As shown in Section~\ref{subsec:iterative-refine} and Figure~\ref{fig:iterative_bar_chart}, performance gains are largest at the second iteration, with diminishing returns thereafter. The complete algorithm is detailed in Algorithm~\ref{alg:instructdiff} (Appendix~\ref{app:algorithm}).

\section{Experiments}

\subsection{Experimental Setup}

\paragraph{Datasets and Models.} 
We evaluate \mname\ across four domains: \textbf{mathematics} using NuminaMath~\citep{numina_math_datasets} (10k samples) with Qwen2.5-7B~\citep{qwen2025qwen25technicalreport}, \textbf{general instruction-following} using Alpaca~\citep{alpaca_eval} (10k samples) with LLaMA3-8B~\citep{dubey2024llama}, \textbf{medical QA} using MedCAQA~\citep{pal2022medmcqa} (10k samples) with LLaMA3-8B, and \textbf{code generation} using BigCode~\citep{cassano2024knowledge} (10k samples) with LLaMA3-8B. Detailed dataset characteristics are in Appendix~\ref{app:dataset_details}.

\paragraph{Baselines.} 

We compare against: (1) \textbf{Random Sampling}; (2) \textbf{PPL-based} (PPL$_{\mathrm{Min}}$, PPL$_{\mathrm{Mid}}$, PPL$_{\mathrm{Max}}$); (3) \textbf{Entropy-based} (Entropy$_{\mathrm{Min}}$, Entropy$_{\mathrm{Mid}}$, Entropy$_{\mathrm{Max}}$); (4) \textbf{Length-based} (Resp Len$_{\mathrm{Max}}$, Inst Len$_{\mathrm{Max}}$, Inst/Resp$_{\mathrm{Max}}$, Inst/Resp$_{\mathrm{Min}}$); (5) \textbf{IFD}~\citep{li2024quantity}, \textbf{Superfiltering}~\citep{li2024superfiltering}, \textbf{SelectIT}~\citep{liu2024selectit}, and \textbf{ZIP}~\citep{yin2024entropy}; (6) \textbf{Full Training} on complete 10k data. Implementation details are in Appendix~\ref{app:baseline_details}.

\paragraph{Training and Evaluation.}
All models use full-parameter supervised fine-tuning. For \mname, key hyperparameters include warmup ratio $\alpha=0.1$, NLL rejection ratio $\gamma=0.1$, and selection ratio $\beta=0.1$ (code: $\beta=0.2$). Evaluation is conducted on standard benchmark splits in each domain. More details on training settings, evaluation protocols, and benchmark breakdowns are provided in Appendix~\ref{app:training_details}.

\begin{table*}[t]
\centering
\small
\setlength{\tabcolsep}{3pt}
\resizebox{\textwidth}{!}{
\begin{tabular}{l | c c c c c c | c c c}
\toprule
 &  \multicolumn{6}{c|}{\textbf{Math Domain (\textit{Qwen2.5-7B / select 10\%})}} & \multicolumn{3}{c}{\textbf{General (\textit{LLaMA3-8B / select 10\%})}} \\
\cmidrule(lr){2-7} \cmidrule(lr){8-10}
\textbf{Method} &
AIME24 & Math-OAI & Minerva & Olympiad & ACM23 & Avg &
AlpacaEval (\%) & Arena-Hard (\%) & Avg \\
\midrule
Base Model &
1.65 & 28.79 & 9.26 & 7.69 & 15.65 & 12.61 &
1.36 & 0.26 & 0.81 \\
\rowcolor{gray!10}
ALL &
5.00 {\color{olive}(+3.35)} & 57.93 {\color{olive}(+29.14)} & 17.83 {\color{olive}(+8.57)} & 22.30 {\color{olive}(+14.61)} & 32.19 {\color{olive}(+16.54)} & 27.05 {\color{olive}(+14.44)} &
8.15 {\color{olive}(+6.79)} & 1.70 {\color{olive}(+1.44)} & 4.93 {\color{olive}(+4.12)} \\

\midrule
PPL$_{\mathrm{Min}}$ &
5.63 {\color{olive}(+3.98)} & 57.80 {\color{olive}(+29.01)} & 16.35 {\color{olive}(+7.09)} & 23.68 {\color{olive}(+13.99)} & 31.72 {\color{olive}(+16.07)} & 27.04 {\color{olive}(+14.43)} &
9.20 {\color{olive}(+7.84)} & 2.12 {\color{olive}(+1.86)} & 5.66 {\color{olive}(+4.85)} \\

PPL$_{\mathrm{Mid}}$ &
6.05 {\color{olive}(+4.40)} & 54.76 {\color{olive}(+25.97)} & 13.61 {\color{olive}(+4.35)} & 21.01 {\color{olive}(+13.32)} & 32.81 {\color{olive}(+17.16)} & 25.65 {\color{olive}(+13.04)} &
7.25 {\color{olive}(+5.89)} & 2.51 {\color{olive}(+2.25)} & 4.88 {\color{olive}(+4.07)} \\

PPL$_{\mathrm{Max}}$ &
3.95 {\color{olive}(+2.30)} & 50.56 {\color{olive}(+21.77)} & 12.77 {\color{olive}(+3.51)} & 18.57 {\color{olive}(+10.88)} & 28.28 {\color{olive}(+12.63)} & 22.83 {\color{olive}(+10.22)} &
9.72 {\color{olive}(+8.36)} & 1.54 {\color{olive}(+1.28)} & 5.63 {\color{olive}(+4.82)} \\

\midrule
Entropy$_{\mathrm{Min}}$ &
5.83 {\color{olive}(+4.18)} & 60.74 {\color{olive}(+31.95)} & 18.59 {\color{olive}(+9.33)} & 27.13 {\color{olive}(+19.44)} & 36.72 {\color{olive}(+21.07)} & 29.80 {\color{olive}(+17.19)} &
10.45 {\color{olive}(+9.09)} & \textbf{3.51 {\color{olive}(+3.25)}} & 6.98 {\color{olive}(+6.17)} \\

Entropy$_{\mathrm{Mid}}$ &
3.54 {\color{olive}(+1.89)} & 54.50 {\color{olive}(+25.71)} & 12.27 {\color{olive}(+3.01)} & 20.12 {\color{olive}(+12.43)} & 29.38 {\color{olive}(+13.73)} & 23.96 {\color{olive}(+11.35)} &
7.44 {\color{olive}(+6.08)} & 2.36 {\color{olive}(+2.10)} & 4.90 {\color{olive}(+4.09)} \\

Entropy$_{\mathrm{Max}}$ &
3.12 {\color{olive}(+1.47)} & 50.50 {\color{olive}(+21.71)} & 12.41 {\color{olive}(+3.15)} & 18.54 {\color{olive}(+10.85)} & 23.59 {\color{olive}(+7.94)} & 21.63 {\color{olive}(+9.02)} &
7.50 {\color{olive}(+6.14)} & 1.53 {\color{olive}(+1.27)} & 4.52 {\color{olive}(+3.71)} \\

\midrule
Resp Len$_{\mathrm{Max}}$ &
2.28 {\color{olive}(+0.63)} & 44.65 {\color{olive}(+15.86)} & 10.02 {\color{olive}(+0.76)} & 15.38 {\color{olive}(+7.69)} & 21.56 {\color{olive}(+5.91)} & 18.78 {\color{olive}(+6.17)} &
11.04 {\color{olive}(+9.68)} & 2.91 {\color{olive}(+2.65)} & 6.98 {\color{olive}(+6.17)} \\

Inst Len$_{\mathrm{Max}}$ &
4.79 {\color{olive}(+3.14)} & 54.98 {\color{olive}(+26.19)} & 12.27 {\color{olive}(+3.01)} & 20.99 {\color{olive}(+13.30)} & 29.53 {\color{olive}(+13.88)} & 24.51 {\color{olive}(+11.90)} &
5.47 {\color{olive}(+4.11)} & 2.02 {\color{olive}(+1.76)} & 3.75 {\color{olive}(+2.94)} \\

Inst/Resp$_{\mathrm{Max}}$ &
3.33 {\color{olive}(+1.68)} & 50.33 {\color{olive}(+21.54)} & 12.08 {\color{olive}(+2.82)} & 16.96 {\color{olive}(+9.27)} & 28.28 {\color{olive}(+12.63)} & 22.20 {\color{olive}(+9.59)} &
4.60 {\color{olive}(+3.24)} & 3.20 {\color{olive}(+1.26)} & 3.90 {\color{olive}(+3.09)} \\

Inst/Resp$_{\mathrm{Min}}$ &
3.74 {\color{olive}(+2.09)} & 52.50 {\color{olive}(+23.71)} & 12.27 {\color{olive}(+3.01)} & 20.14 {\color{olive}(+12.45)} & 27.97 {\color{olive}(+12.32)} & 23.32 {\color{olive}(+10.71)} &
9.44 {\color{olive}(+8.08)} & 2.74 {\color{olive}(+2.38)} & 6.09 {\color{olive}(+5.28)} \\

\midrule
Random &
5.43 {\color{olive}(+3.78)} & 55.41 {\color{olive}(+26.62)} & 14.86 {\color{olive}(+5.60)} & 20.18 {\color{olive}(+12.49)} & 33.13 {\color{olive}(+17.48)} & 25.80 {\color{olive}(+13.19)} &
5.44 {\color{olive}(+4.08)} & 2.18 {\color{olive}(+1.92)} & 3.81 {\color{olive}(+3.00)} \\

IFD &
3.12 {\color{olive}(+1.47)} & 50.36 {\color{olive}(+21.57)} & 13.98 {\color{olive}(+4.72)} & 18.14 {\color{olive}(+10.45)} & 24.38 {\color{olive}(+8.73)} & 21.99 {\color{olive}(+9.38)} &
10.75 {\color{olive}(+9.39)} & 2.29 {\color{olive}(+2.03)} & 6.52 {\color{olive}(+5.71)} \\

SelectIT &
4.15 {\color{olive}(+2.50)} & 52.86 {\color{olive}(+24.07)} & 15.06 {\color{olive}(+5.80)} & 20.29 {\color{olive}(+12.60)} & 29.38 {\color{olive}(+13.73)} & 24.35 {\color{olive}(+11.74)} &
7.84 {\color{olive}(+6.48)} & 2.34 {\color{olive}(+2.08)} & 5.09 {\color{olive}(+4.28)} \\

ZIP &
4.15 {\color{olive}(+2.50)} & 50.30 {\color{olive}(+21.51)} & 12.23 {\color{olive}(+2.97)} & 16.69 {\color{olive}(+8.99)} & 30.00 {\color{olive}(+14.35)} & 22.61 {\color{olive}(+10.00)} &
6.90 {\color{olive}(+5.54)} & 0.78 {\color{olive}(+0.52)} & 3.84 {\color{olive}(+3.03)} \\

Superfiltering &
4.80 {\color{olive}(+3.15)} & 55.55 {\color{olive}(+26.76)} & 14.06 {\color{olive}(+4.80)} & 21.38 {\color{olive}(+13.69)} & 30.78 {\color{olive}(+15.13)} & 25.31 {\color{olive}(+12.70)} &
12.08 {\color{olive}(+10.72)} & 2.61 {\color{olive}(+2.35)} & 7.35 {\color{olive}(+6.54)} \\

\midrule
\rowcolor{blue!10}\textbf{Ours} &
\textbf{7.71 {\color{olive}(+6.06)}} & \textbf{61.79 {\color{olive}(+33.00)}} & \textbf{21.42 {\color{olive}(+12.16)}} & \textbf{26.94 {\color{olive}(+19.25)}} & \textbf{40.31 {\color{olive}(+24.66)}} & \textbf{31.63 {\color{olive}(+19.02)}} &
\textbf{12.09 {\color{olive}(+10.73)}} & 2.84 {\color{olive}(+2.58)} & \textbf{7.47 {\color{olive}(+6.66)}} \\

\bottomrule
\end{tabular}
}
\caption{Comparison of data selection strategies across Math and General domains. 
Values are annotated with the change relative to the Base Model (red for decrease, olive for increase). 
Math benchmarks are evaluated with Qwen2.5-7B, while General benchmarks are evaluated with LLaMA3-8B.}
\label{tab:main_results_math_general}
\end{table*}

\subsection{Main Results}

Tables~\ref{tab:main_results_math_general} and~\ref{tab:main_results_med_code} present our main results across four domains. \mname\ consistently achieves the best performance using only 10\% of training data (20\% for code), outperforming all baselines including full-data training. \mname\ is the \textit{only} method that consistently surpasses full training performance while using merely 10-20\% of data, achieving average relative improvements of +17\% (math), +6.2\% (medical), +52\% (general), and +4.9\% (code) over full training. On mathematics (Qwen2.5-7B), \mname\ achieves 31.63 average score versus 27.05 for full training, with particularly strong gains on AIME 2024 (7.71 vs. 5.00, +54\%). For medical QA (LLaMA3-8B), \mname\ reaches 56.42 average accuracy compared to 53.14 for full training (+6.2\%). On general instruction-following (LLaMA3-8B), \mname\ achieves 12.09\% LC win rate versus 8.15\% for full training (+48\%). For code generation (LLaMA3-8B, 20\% data), \mname\ scores 45.1 versus 43.0 for full training (+4.9\%). These results validate that entropy-guided selection identifies samples at the model's learnable frontier, enabling more efficient and effective fine-tuning than volume-based approaches.

\begin{table*}[t]
\centering
\small
\setlength{\tabcolsep}{3pt}
\resizebox{\textwidth}{!}{
\begin{tabular}{l | c c c c | c c c c c c}
\toprule
 & \multicolumn{4}{c|}{\textbf{Medical (\textit{LLaMA3-8B / select 10\%}) }} & \multicolumn{6}{c}{\textbf{Code (\textit{LLaMA3-8B / select 20\%})}} \\
\cmidrule(lr){2-5} \cmidrule(lr){6-11}
\textbf{Method} &
MedQA & MMLU & MedMCQA & Avg &
HumanEval & HumanEval+ & MBPP & MBPP+ & Bigcode & Avg \\
\midrule
Base Model &
43.60 & 51.04 & 48.98 & 47.87 &
40.2 & 33.5 & 55.3 & 46.3 & 22.5 & 39.6 \\
\rowcolor{gray!10}
ALL &
49.96 {\color{olive}(+6.36)} & 61.04 {\color{olive}(+10.00)} & 48.41 {\color{red}(-0.57)} & 53.14 {\color{olive}(+5.27)} &
44.5 {\color{olive}(+4.3)} & 35.4 {\color{olive}(+1.9)} & 59.8 {\color{olive}(+4.5)} & 49.7 {\color{olive}(+3.4)} & \textbf{25.8 {\color{olive}(+3.3)}} & 43.0 {\color{olive}(+3.4)}\\

\midrule
PPL$_{\mathrm{Min}}$ &
48.23 {\color{olive}(+4.63)} & 57.69 {\color{olive}(+6.65)} & 43.51 {\color{red}(-5.47)} & 49.81 {\color{olive}(+1.94)} &
41.5 {\color{olive}(+1.3)} & 34.1 {\color{olive}(+0.6)} & 56.1 {\color{olive}(+0.8)} & 46.3 {\color{red}(-0.0)} & 23.4 {\color{olive}(+0.9)} & 40.3 {\color{olive}(+0.7)} \\

PPL$_{\mathrm{Mid}}$ &
43.75 {\color{red}(-0.85)} & 48.45 {\color{red}(-2.59)} & 35.09 {\color{red}(-13.89)} & 42.43 {\color{red}(-5.44)} &
44.5 {\color{olive}(+4.3)} & 36.6 {\color{olive}(+3.1)} & 56.9 {\color{olive}(+1.6)} & 46.8 {\color{red}(-0.5)} & 21.6 {\color{red}(-0.9)} & 41.3 {\color{olive}(+1.7)} \\

PPL$_{\mathrm{Max}}$ &
47.76 {\color{olive}(+4.16)} & 58.38 {\color{olive}(+7.34)} & 45.37 {\color{red}(-3.61)} & 50.50 {\color{olive}(+2.63)} &
40.2 {\color{red}(-0.0)} & 34.1 {\color{olive}(+0.6)} & 55.6 {\color{olive}(+0.3)} & 47.4 {\color{olive}(+1.1)} & 21.0 {\color{red}(-1.5)} & 39.7 {\color{red}(-0.1)} \\

\midrule
Entropy$_{\mathrm{Min}}$ &
36.76 {\color{red}(-6.84)} & 42.24 {\color{red}(-8.80)} & 34.23 {\color{red}(-14.75)} & 37.74 {\color{red}(-10.13)} &
46.3 {\color{olive}(+6.1)} & 39.0 {\color{olive}(+5.5)} & 58.7 {\color{olive}(+3.4)} & 46.3 {\color{red}(-0.0)} & 23.4 {\color{olive}(+0.9)} & 42.7 {\color{olive}(+3.1)} \\

Entropy$_{\mathrm{Mid}}$ &
46.43 {\color{olive}(+2.83)} & 53.30 {\color{olive}(+2.26)} & 42.19 {\color{red}(-6.79)} & 47.31 {\color{red}(-0.56)} &
41.5 {\color{olive}(+1.3)} & 34.1 {\color{olive}(+0.6)} & 59.3 {\color{olive}(+4.0)} & 48.1 {\color{olive}(+1.8)} & 23.8 {\color{olive}(+1.3)} & 41.4 {\color{olive}(+1.8)} \\

Entropy$_{\mathrm{Max}}$ &
52.40 {\color{olive}(+8.80)} & \textbf{66.52 {\color{olive}(+15.48)}} & \textbf{52.16 {\color{olive}(+3.18)}} & \textbf{57.03 {\color{olive}(+9.16)}} &
40.9 {\color{olive}(+0.7)} & 34.8 {\color{olive}(+1.3)} & 57.9 {\color{olive}(+2.6)} & 48.4 {\color{olive}(+2.1)} & 21.1 {\color{red}(-1.4)} & 40.6 {\color{olive}(+1.0)} \\

\midrule
Resp Len$_{\mathrm{Max}}$ &
42.11 {\color{red}(-1.49)} & 41.99 {\color{red}(-9.05)} & 32.44 {\color{red}(-16.54)} & 38.85 {\color{red}(-9.02)} &
47.0 {\color{olive}(+6.8)} & 37.8 {\color{olive}(+4.3)} & 57.4 {\color{olive}(+2.1)} & 48.4 {\color{olive}(+2.1)} & 24.2 {\color{olive}(+1.7)} & 43.0 {\color{olive}(+3.4)} \\

Inst Len$_{\mathrm{Max}}$ &
49.25 {\color{olive}(+5.65)} & 57.10 {\color{olive}(+6.06)} & 45.52 {\color{red}(-3.46)} & 50.62 {\color{olive}(+2.75)} &
42.1 {\color{olive}(+1.9)} & 35.4 {\color{olive}(+1.9)} & 57.7 {\color{olive}(+2.4)} & 46.3 {\color{red}(-0.0)} & 23.6 {\color{olive}(+1.1)} & 41.0 {\color{olive}(+1.4)} \\

Inst/Resp$_{\mathrm{Max}}$ &
50.98 {\color{olive}(+7.38)} & 58.71 {\color{olive}(+7.67)} & 43.92 {\color{red}(-5.06)} & 51.20 {\color{olive}(+3.33)} &
43.9 {\color{olive}(+3.7)} & 38.4 {\color{olive}(+4.9)} & 59.0 {\color{olive}(+3.7)} & 49.2 {\color{olive}(+2.9)} & 21.9 {\color{red}(-0.6)} & 42.5 {\color{olive}(+2.9)} \\

Inst/Resp$_{\mathrm{Min}}$ &
35.51 {\color{red}(-8.09)} & 44.65 {\color{red}(-6.39)} & 40.16 {\color{red}(-8.82)} & 40.11 {\color{red}(-7.76)} &
47.0 {\color{olive}(+6.8)} & 38.4 {\color{olive}(+4.9)} & 56.1 {\color{olive}(+0.8)} & 46.3 {\color{red}(-0.0)} & 22.8 {\color{olive}(+0.3)} & 42.1 {\color{olive}(+2.5)} \\

\midrule
Random &
45.95 {\color{olive}(+2.35)} & 54.22 {\color{olive}(+3.18)} & 40.11 {\color{red}(-8.87)} & 46.76 {\color{red}(-1.11)} &
42.1 {\color{olive}(+1.9)} & 33.5 {\color{red}(-0.0)} & 57.7 {\color{olive}(+2.4)} & 47.9 {\color{olive}(+1.6)} & 22.9 {\color{olive}(+0.4)} & 40.8 {\color{olive}(+1.2)} \\

IFD &
43.13 {\color{red}(-0.47)} & 48.27 {\color{red}(-2.77)} & 42.36 {\color{red}(-6.62)} & 44.59 {\color{red}(-3.28)} &
40.9 {\color{olive}(+0.7)} & 34.1 {\color{olive}(+0.6)} & 51.6 {\color{red}(-3.7)} & 40.7 {\color{red}(-5.6)} & 18.2 {\color{red}(-4.3)} & 37.1 {\color{red}(-2.5)} \\

SelectIT &
46.11 {\color{olive}(+2.51)} & 46.84 {\color{red}(-4.20)} & 35.93 {\color{red}(-13.05)} & 42.96 {\color{red}(-4.91)} &
48.2 {\color{olive}(+8.0)} & 38.4 {\color{olive}(+4.9)} & 60.6 {\color{olive}(+5.3)} & \textbf{51.9 {\color{olive}(+5.6)}} & 23.0 {\color{olive}(+0.5)} & 44.4 {\color{olive}(+4.8)} \\

ZIP &
51.53 {\color{olive}(+7.93)} & 62.87 {\color{olive}(+11.83)} & 48.84 {\color{red}(-0.14)} & 54.41 {\color{olive}(+6.54)} &
38.4 {\color{red}(-1.8)} & 32.9 {\color{red}(-0.6)} & 59.3 {\color{olive}(+4.0)} & 50.3 {\color{olive}(+4.0)} & 16.0 {\color{red}(-6.5)} & 39.4 {\color{red}(-0.2)} \\

Superfiltering &
41.63 {\color{red}(-1.97)} & 41.15 {\color{red}(-9.89)} & 32.08 {\color{red}(-16.90)} & 38.29 {\color{red}(-9.58)} &
43.3 {\color{olive}(+3.1)} & 34.8 {\color{olive}(+1.3)} & 54.2 {\color{red}(-1.1)} & 44.2 {\color{red}(-2.1)} & 20.6 {\color{red}(-1.9)} & 39.4 {\color{red}(-0.2)} \\

\midrule
\rowcolor{blue!10}\textbf{Ours} &
\textbf{54.67 {\color{olive}(+11.07)}} & 64.48 {\color{olive}(+13.44)} & 50.11 {\color{olive}(+1.13)} & 56.42 {\color{olive}(+8.55)} &
\textbf{48.2 {\color{olive}(+8.0)}} & \textbf{40.9 {\color{olive}(+7.4)}} &\textbf{ 60.6 {\color{olive}(+5.3)}} & 50.0 {\color{olive}(+3.7)} & 25.7 {\color{olive}(+3.2)} & \textbf{45.1 {\color{olive}(+5.5)}} \\

\bottomrule
\end{tabular}
}
\caption{Comparison of data selection strategies across Medical and Code domains, with relative changes from Base Model shown in {\color{olive}olive} (+) and {\color{red}red} (-). Both domains use LLaMA3-8B. Avg is the average across the metrics of each domain.}
\label{tab:main_results_med_code}
\end{table*}

\subsection{Ablation Studies}
\label{subsec:ablation}

We conduct comprehensive ablation studies to validate each component of \mname, covering: (1) bi-directional NLL filtering, (2) entropy selection range and (3) warmup data size.

\paragraph{Bi-Directional NLL Filtering.}
Table~\ref{tab:ablation_nll} demonstrates that removing bi-directional filtering causes substantial performance drops: 2.59 points on mathematics (31.69 → 29.10), 1.66 points on general instruction-following (12.09 → 10.43), and 5.56 points on medical QA (56.42 → 50.86). This validates that extreme $\Delta$NLL samples, representing either redundant knowledge (bottom percentile) or incomprehensible patterns (top percentile), introduce training noise. The filtering mechanism effectively identifies the learnable frontier where samples provide meaningful gradient signals without catastrophic forgetting risks.

\begin{table}[h]
\centering
\small

\begin{tabular}{lcccc}
\toprule
\textbf{Method} & \textbf{Math} & \textbf{General} & \textbf{Medical} & \textbf{Code} \\
\midrule
Full pipeline & 31.69 & 12.09 & 56.42 & 45.1 \\
w/o NLL filter & 29.10 & 10.43 & 50.86 & 40.2 \\
\midrule
$\Delta$ & {\color{red}-2.59} & {\color{red}-1.66} & {\color{red}-5.56} & {\color{red}-4.9} \\
\bottomrule
\end{tabular}
\vspace{-2mm}
\caption{\textbf{Ablation: Bi-Directional NLL Filtering.} Removing NLL filtering degrades performance across all domains.}
\label{tab:ablation_nll}
\end{table}

\begin{figure}[t]
    \centering
    \includegraphics[width=0.95\columnwidth]{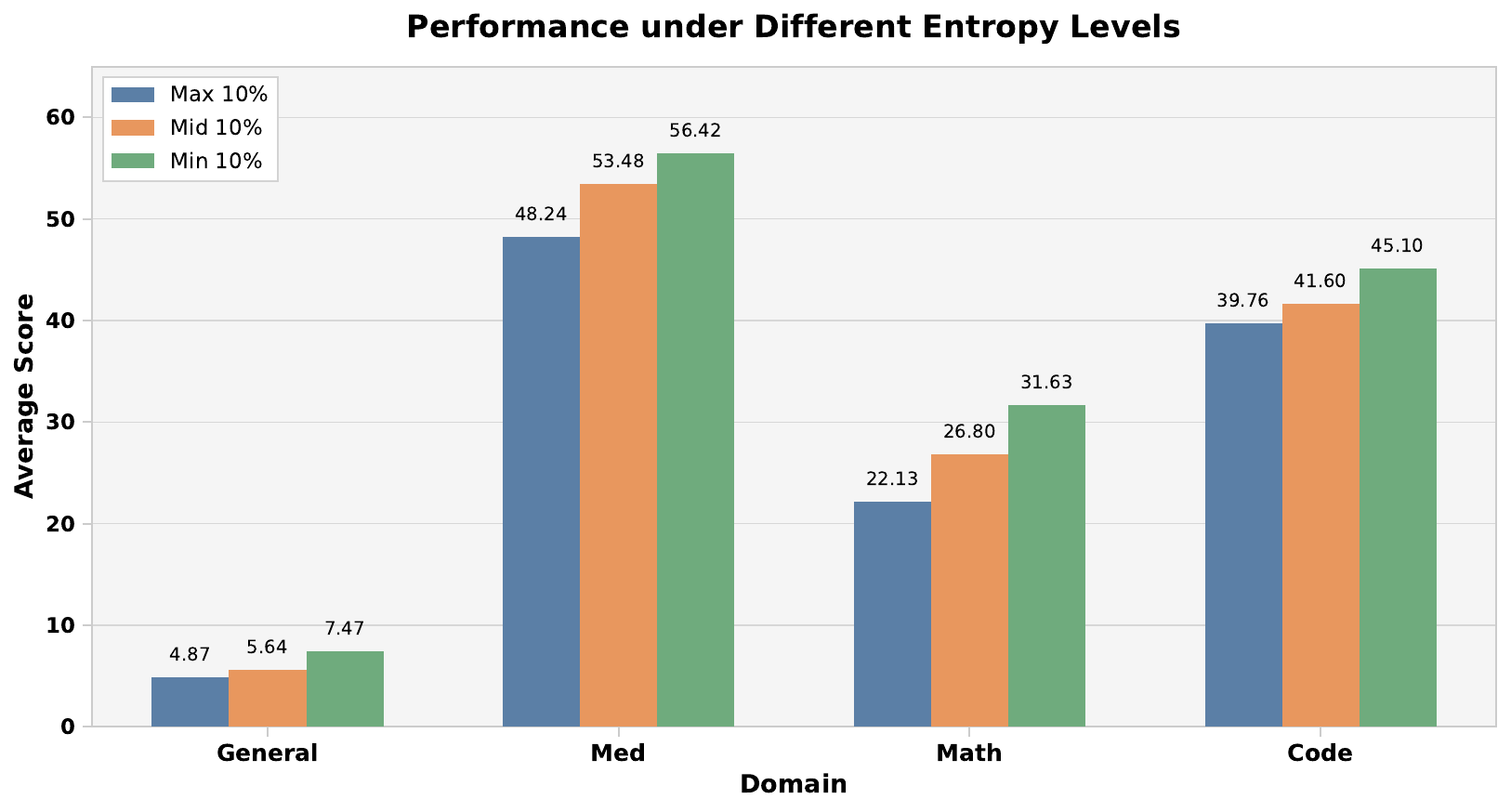}
    \caption{\textbf{Entropy Selection Range Analysis.} Bottom 10\% (low $\Delta H$) consistently outperforms, validating domain-adaptive entropy dynamics.}
    \label{fig:entropy_range}
    \vspace{-2mm}
\end{figure}

\paragraph{Entropy Selection Range.}
We systematically evaluate three entropy selection strategies: top 10\% (Max 10\%, highest $\Delta H$), middle 10\% (Mid 10\%), and bottom 10\% (Min 10\%, lowest $\Delta H$). Figure~\ref{fig:entropy_range} shows that bottom 10\% (low $\Delta H$) consistently outperforms across all domains. For mathematics, Min 10\% achieves 31.63 versus 22.13 for Max 10\% (+43\% relative gain). For general instruction-following, Min 10\% reaches 7.47 versus 4.87 for Max 10\% (+53\%). Medical and code domains show similar patterns: Min 10\% achieves 56.42 (vs. 48.24 Max) and 45.19 (vs. 39.76 Max) respectively. Middle-range selections consistently underperform across all domains. These results validate our core principle: selecting samples with the lowest differential entropy consistently yields superior performance.

\paragraph{Warmup Data Size.}
Table~\ref{tab:warmup_size} analyzes warmup set size trade-offs. We evaluate four ratios: 1\%, 10\%, 20\%, and 50\%. Results show 10\% (1k samples) provides optimal balance, achieving 31.63 average score versus 25.03 for 1\% and 29.59 for 50\%. The 1\% degradation stems from insufficient calibration—only 100 samples cannot capture representative instruction-tuning dynamics, yielding noisy gap estimates. The 50\% decline indicates diminishing returns: larger warmup sets consume excessive resources without improving distributional measurements. The 10\% setting achieves strong performance across all benchmarks (AIME 2024: 7.71, Olympiad: 26.94), and even 20\% warmup (31.22) performs comparably, suggesting robustness to moderate variations.

\begin{table}[t]
\centering

\resizebox{1.0\columnwidth}{!}{
\begin{tabular}{ccccccc}
\toprule
\textbf{Warmup} & \textbf{AIME24} & \textbf{Math-OAI} & \textbf{Minerva} & \textbf{Olympiad} & \textbf{ACM23} & \textbf{Avg} \\
\midrule
1\% & 3.32 & 55.61 & 14.91 & 21.13 & 30.16 & 25.03 \\
10\% & \textbf{7.71} & \textbf{61.79} & 21.42 & \textbf{26.94} & \textbf{40.31} & \textbf{31.63} \\
20\% & 7.51 & 61.36 & \textbf{23.35} & 26.98 & 36.88 & 31.22 \\
50\% & 6.25 & 60.68 & 21.14 & 25.81 & 34.06 & 29.59 \\
\bottomrule
\end{tabular}}
\caption{\textbf{Ablation: Warmup Data Size.} 10\% warmup provides optimal balance between calibration quality and computational cost.}
\label{tab:warmup_size}
\vspace{-2mm}
\end{table}

\section{Analyses}
\label{sec:analyses}

\subsection{Understanding Entropy-Guided Selection}
\label{subsec:entropy-guided}

\paragraph{Why Low $\Delta H$ Works Across Domains.}
To understand the mechanism behind $\Delta H$'s effectiveness, we analyze its correlation with various metrics in Figure~\ref{fig:correlation} (detailed metric definitions in Appendix~\ref{app:metrics}). $\Delta H$ exhibits moderate correlation with $\Delta$NLL but negligible correlation with length, confirming it captures cognitive alignment rather than superficial statistics. The effectiveness stems from two key insights. First, \textbf{model-awareness}: optimal samples must be evaluated relative to the base model's current knowledge state, not in isolation. Second, \textbf{domain-adaptive learning trajectories}: the manifestation of low $\Delta H$ fundamentally differs by task structure. For mathematical reasoning, low $\Delta H$ corresponds to \textit{negative} values—entropy increase during calibration. These samples require the model to expand solution path diversity, as problem-solving benefits from exploratory reasoning~\cite{cheng2025reasoning}. The base model already possesses relevant knowledge~\cite{zhou2024lima}; effective fine-tuning activates diverse reasoning strategies. Conversely, for general instruction-following and medical QA, low $\Delta H$ manifests as \textit{small positive} values—entropy decrease. Here, fine-tuning compresses diffuse priors onto canonical behaviors~\cite{yin2024entropy}, representing knowledge activation rather than injection. Weak correlations with complexity proxies further validate that gains arise from aligning with learning dynamics rather than task difficulty alone. Unlike single-model metrics, $\Delta H$ directly measures distributional shifts, automatically identifying samples at the model's learnable frontier across domains.

\begin{figure}[t]
    \centering
    \includegraphics[width=0.90\columnwidth]{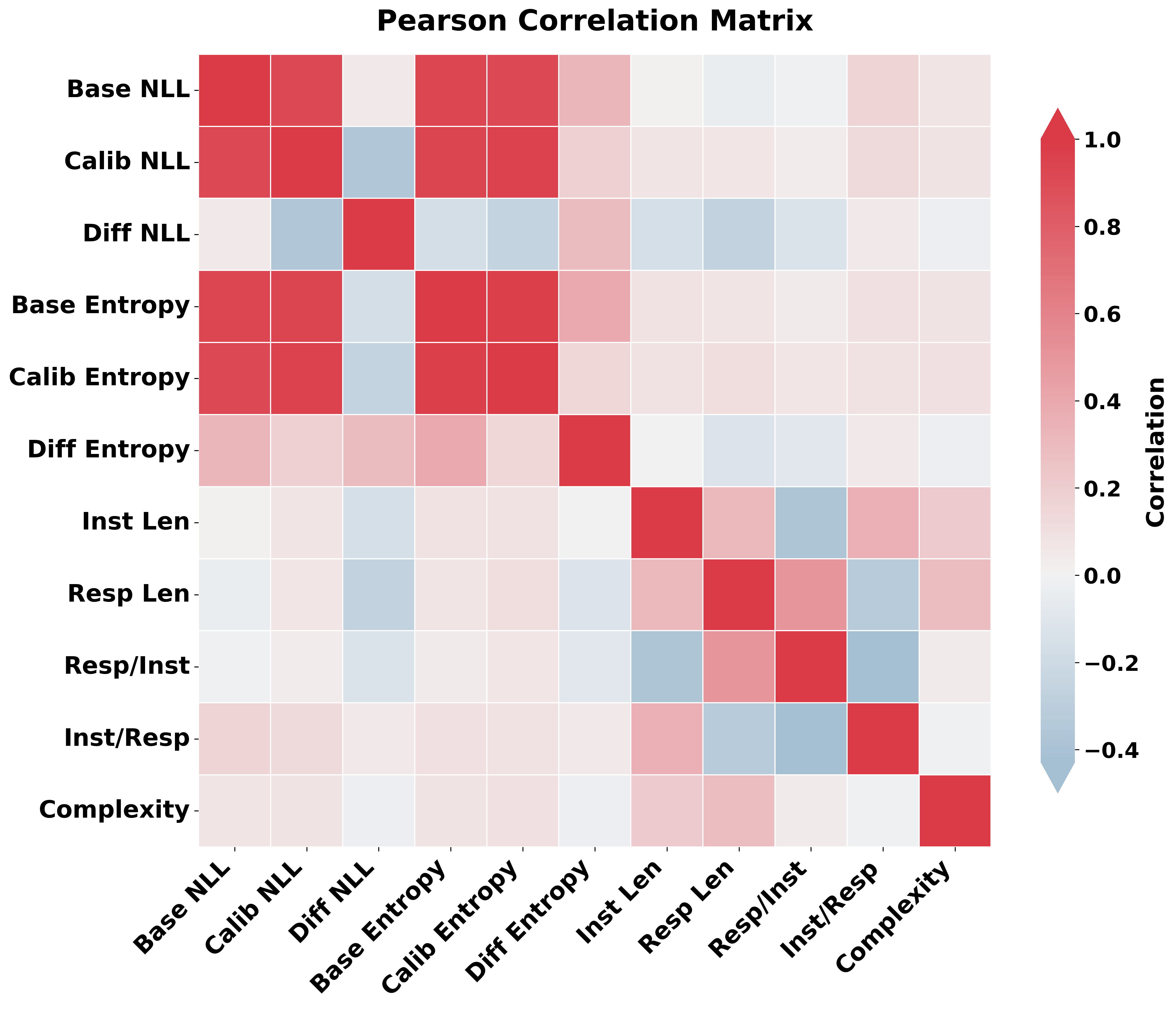}
    \caption{\textbf{Correlation Matrix.} $\Delta H$ measures cognitive alignment orthogonal to difficulty metrics.}
    \label{fig:correlation}
\end{figure}

\subsection{Scaling to Large-Scale Data Selection}
\label{subsec:scaling}

We evaluate \mname's effectiveness on larger datasets by selecting 10k from 100k samples for both mathematics and general instruction-following, comparing against random selection and full training.
Table~\ref{tab:scaling} shows \mname\ successfully scales to large pools. On mathematics, selecting 10\% from 100k achieves 29.80 average score, outperforming random-10\% (27.94, +6.7\% relative) and full-100k training (27.16, +9.7\% relative). For general instruction-following, \mname\ achieves 8.01 average score versus 7.20 for random (+11.3\%) and 7.34 for full training (+9.1\%). These results indicate entropy-guided principles generalize effectively to larger data pools, with relative gains maintained even at scale.

\begin{table}[t]
\centering

\small

\begin{tabular}{lcccc}
\toprule
\textbf{Domain} & \textbf{Method} & \textbf{Pool} & \textbf{Selected} & \textbf{Avg} \\
\midrule
\multirow{3}{*}{Math} 
& Full & 100k & 100\%  &  27.16 \\
& Random & 100k & 10\% &  27.94 \\
& Superfilter & 100k & 10\% & 27.51 \\
& \textbf{Ours} & 100k & 10\% & \textbf{29.80}\\
\midrule
\multirow{3}{*}{General} 
& Full & 52k & 10\%  & 7.34 \\
& Random & 52k & 10\%  & 7.20 \\
& Superfilter & 52k & 10\%  & 6.61 \\
& \textbf{Ours} & 52k & 10\%  & \textbf{8.01} \\
\bottomrule
\end{tabular}
\vspace{-2mm}
\caption{\textbf{Scaling to Large-Scale Selection.} \mname\ maintains effectiveness at scale, outperforming both random selection and full training.}
\label{tab:scaling}
\end{table}

\begin{tcolorbox}[takeawaystyle, title={\textcolor{goldenstar}{\faStar}~\textbf{Takeaway 1:}}]
\mname\ scales to 100k data pools, where selecting 10\% outperforms both full training (+9.7\%) and random selection (+6.7\%), with efficiency gains amplifying at larger scales.
\end{tcolorbox}

\subsection{Weak-to-Strong Calibration for Efficient Selection}
\label{subsec:weak-to-strong}

We investigate whether smaller models can calibrate data selection for larger models by using Qwen2.5-0.5B and Qwen2.5-1.5B to select data for training Qwen2.5-7B. Table~\ref{tab:weak2strong} shows weak-to-strong calibration achieves competitive performance. Qwen2.5-7B calibrated using Qwen2.5-0.5B achieves 28.4 average score, +10.1\% over random (25.80) and only 10.2\% below same-size 7B calibration (31.63). Qwen2.5-1.5B calibration reaches 27.6 average, +7.0\% over random, offering a middle ground between efficiency and performance. The computational advantage is substantial: 0.5B or 1.5B calibration reduces warmup computation by up to 14$\times$ or 4.7$\times$ respectively based on parameter count ratios, while retaining most performance gains, providing an efficient option for compute-constrained practitioners.

\begin{table}[t]
\centering

\resizebox{1.0\columnwidth}{!}{
\begin{tabular}{lccccccc}
\toprule
\textbf{Calibration} & \textbf{Target} & \textbf{AIME24} & \textbf{Math-OAI} & \textbf{Minerva} & \textbf{Olympiad} & \textbf{ACM23} & \textbf{Avg} \\
\midrule
Random & Qwen2.5-7B & 5.43 & 55.41 & 14.86 & 20.18 & 33.13 & 25.80 \\
\midrule
Qwen2.5-0.5B & Qwen2.5-7B & 6.67 & 59.35 & 16.76 & 23.91 & 35.31 & 28.4 \\
Qwen2.5-1.5B & Qwen2.5-7B & 6.03 & 59.5 & 16.24 & 23.71 & 32.5 & 27.6 \\
Qwen2.5-7B & Qwen2.5-7B & \textbf{7.71} & \textbf{61.79} & \textbf{21.42} & \textbf{26.94} & \textbf{40.31} & \textbf{31.63} \\
\bottomrule
\end{tabular}}
\caption{\textbf{Weak-to-Strong Calibration.} Smaller models effectively calibrate for larger models within the same family, enabling 4.2× computational savings.}
\label{tab:weak2strong}
\vspace{-2mm}
\end{table}

\begin{tcolorbox}[takeawaystyle, title={\textcolor{goldenstar}{\faStar}~\textbf{Takeaway 2:}}]
Weak-to-strong calibration using 0.5B model enables 14× computational savings while retaining 89.8\% of same-size calibration performance within the same model family.
\end{tcolorbox}

\subsection{Iterative Refinement Analysis}
\label{subsec:iterative-refine}

\begin{figure}[t]
    \centering
    \includegraphics[width=0.98\columnwidth]{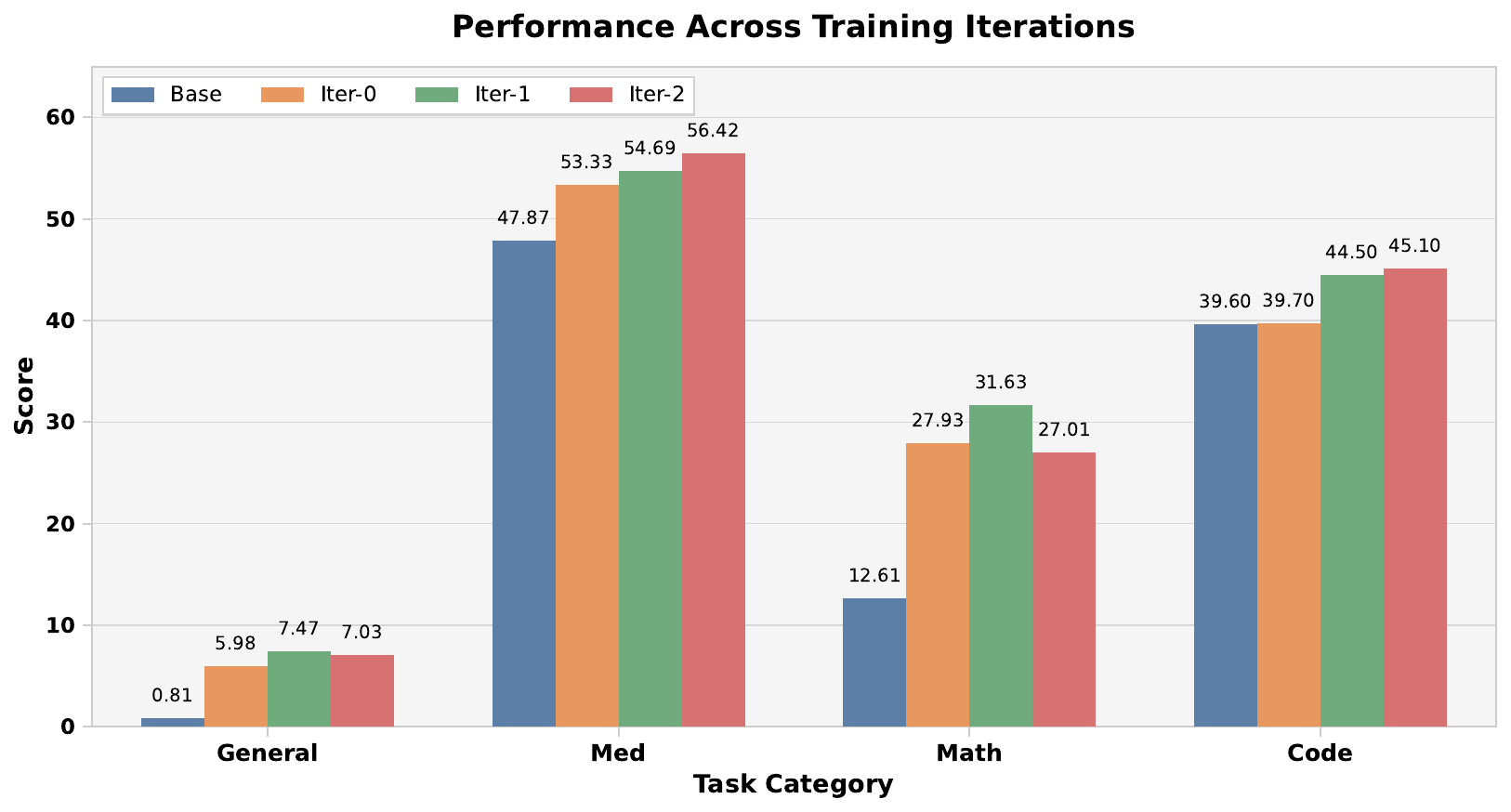}
    \caption{\textbf{Iterative Refinement.} Two iterations provide optimal cost-benefit trade-off before diminishing returns.}
    \label{fig:iterative_bar_chart}
\end{figure}

We assess iterative refinement over up to three iterations across all four domains, using the previously fine-tuned model for calibration each time (Section~\ref{sec:iterative-refinement}). Figure~\ref{fig:iterative_bar_chart} shows the largest gains at the second iteration—for instance, math improves from 31.63 (iter 1) to 32.89 (iter 2, +4.0\%), with diminishing returns at iter 3 (33.24, +1.1\%). Similar trends occur in other domains, suggesting two iterations offer the best balance of performance and cost.

\begin{tcolorbox}[takeawaystyle, title={\textcolor{goldenstar}{\faStar}~\textbf{Takeaway 3:}}]
Iterative refinement improves selection quality consistently, with 2 iterations optimal before diminishing returns emerge.
\end{tcolorbox}

\subsection{Cross-Model Selection Consistency}
\label{subsec:cross-model}
We investigate cross-model selection consistency via two experiments. First, we use different models (Qwen2.5-7B/14B, LLaMA3-8B/3.1-8B) to independently select data from the same pool using \mname, then measure selection overlap (Table 7). Same-family models show 60–64\% overlap, compared to just 15–20\% for cross-family pairs. Second, we train the target model Qwen2.5-7B on data selected by different calibration models (Table 8). Same-family calibration (Qwen2.5-14B) achieves a 30.84 average score, outperforming cross-family calibration (LLaMA: 25.74–26.31) by 17–20\%. These results show that selection consistency is family-dependent: same-family models agree on "good data" and yield better downstream performance.

\begin{table}[t]
\centering
\small
\begin{tabular}{@{}lrrr@{}}
\toprule
\textbf{Model Pair} & \textbf{Overlap} & \textbf{R(A)} & \textbf{R(B)} \\
\midrule
Qwen2.5-7B / Qwen2.5-14B & 545 & 0.64 & 0.62 \\
Qwen2.5-7B / LLaMA3-8B   & 131 & 0.15 & 0.15 \\
Qwen2.5-7B / LLaMA3.1-8B & 164 & 0.19 & 0.15 \\
Qwen2.5-14B / LLaMA3-8B  & 152 & 0.17 & 0.18 \\
Qwen2.5-14B / LLaMA3.1-8B& 178 & 0.20 & 0.17 \\
LLaMA3-8B / LLaMA3.1-8B  & 396 & 0.46 & 0.37 \\
\bottomrule
\end{tabular}
\caption{\textbf{Cross-Model Selection Overlap.} Same-family models show 3-4× higher overlap than cross-family pairs.}
\label{tab:cross_model}
\end{table}

\begin{table}[t]
\centering
\small
\begin{tabular}{lccc}
\toprule
\textbf{Family} & \textbf{Calibration Model} & \textbf{Size} & \textbf{Avg} \\
\midrule
Cross-family & LLaMA3-8B   & 10\% & 26.31 \\
Cross-family & LLaMA3.1-8B & 10\% & 25.74 \\
\midrule
Same-family  & Qwen2.5-7B  & 10\% & 27.93 \\
Same-family  & Qwen2.5-14B & 10\% & \textbf{30.84} \\
\bottomrule
\end{tabular}
\caption{\textbf{Calibration Family Impact.} Target: Qwen2.5-7B. Same-family calibration yields +19.8\% gain over cross-family.}
\label{tab:source_comparison}
\end{table}

\begin{tcolorbox}[takeawaystyle, title={\textcolor{goldenstar}{\faStar}~\textbf{Takeaway 4:}}]
Selection consistency is family-dependent: same-family models share 60-64\% selections versus 15-20\% cross-family, validating the need for matched calibration.
\end{tcolorbox}

\section{Conclusion}
We propose \mname, a simple, unified data selection framework based on differential entropy. Comparing base and calibrated models, we show that selecting examples with lowest differential entropy consistently yields optimal fine-tuning across domains, without expert heuristics. \mname achieves up to 52\% improvement over full-data training using just 10\%–20\% of data and scales efficiently to large datasets with fast calibration. These results show differential entropy is a principled, domain-agnostic criterion for efficient, effective LLM adaptation.

\clearpage
\newpage 

\section*{Limitations}

\textbf{Domain-Specific Tuning.} Our bi-directional reject ratio $\gamma$ and selection ratio $\beta$ require minor adjustments across domains: mathematics benefits from $\gamma = 0.1$, while medical data performs better with $\gamma = 0.05$. Although these variations are modest, fully automated hyperparameter selection remains an open direction for future work.

\noindent\textbf{Warmup Dependency.} Selection quality depends on the calibration model quality. Very small warmup sets (e.g., 1\% or 100 samples) produce noisy gap measurements, as shown in Table~\ref{tab:warmup_size}. We recommend using almost 1,000 warmup samples to ensure stable and reliable entropy difference estimates.

\noindent\textbf{Model Family Specificity.} Cross-model overlap analysis (Table~\ref{tab:cross_model}) indicates that optimal selected subsets vary significantly across model families (same-family overlap: 60-64\% vs. cross-family: 15-20\%). This suggests practitioners should ideally compute distributional gaps using their target base model or a same-family calibration model to maximize selection quality and downstream performance.

\bibliography{custom}

\newpage

\appendix

\section{Dataset Details}
\label{app:dataset_details}

\paragraph{Mathematics (NuminaMath).} The NuminaMath dataset contains 10,000 mathematical reasoning problems covering arithmetic, algebra, geometry, and competition-level mathematics. Problems range from elementary school to Mathematical Olympiad difficulty. Each sample consists of a problem statement (instruction) and a step-by-step solution (response).

\paragraph{General Instruction-Following (Alpaca).} The Alpaca dataset contains 10,000 instruction-following examples spanning diverse tasks including question answering, creative writing, summarization, and brainstorming. Instructions vary in complexity and response length, representing typical user interactions with instruction-tuned models.

\paragraph{Medical QA (MedCAQA).} We randomly sample 10,000 training examples from the MedCAQA dataset, which contains multiple-choice questions from Indian medical entrance exams. Questions cover anatomy, physiology, pharmacology, pathology, and clinical scenarios. Each sample includes a medical question and a detailed explanation.

\paragraph{Code Generation (BigCode).} We randomly sample 10,000 code generation examples from the BigCode dataset. Samples include function signatures with natural language descriptions and corresponding Python implementations, covering algorithms, data structures, and practical programming tasks.

\section{Baseline Implementation Details}
\label{app:baseline_details}

\paragraph{PPL-based Selection.} We compute the per-token perplexity using the base model for all samples, then select the top 10\% based on three strategies: PPL$_{\mathrm{Min}}$ selects samples with lowest perplexity (easiest for base model), PPL$_{\mathrm{Mid}}$ selects samples in the middle 10\% range, and PPL$_{\mathrm{Max}}$ selects samples with highest perplexity (hardest for base model).

\paragraph{Entropy-based Selection.} We compute the average per-token entropy using the base model, then select samples based on three strategies: Entropy$_{\mathrm{Min}}$ (lowest entropy, highest confidence), Entropy$_{\mathrm{Mid}}$ (middle range), and Entropy$_{\mathrm{Max}}$ (highest entropy, lowest confidence).

\paragraph{Length-based Selection.} We implement four variants: Resp Len$_{\mathrm{Max}}$ selects samples with longest responses, Inst Len$_{\mathrm{Max}}$ selects samples with longest instructions, Inst/Resp$_{\mathrm{Max}}$ selects samples with highest instruction-to-response length ratio, and Inst/Resp$_{\mathrm{Min}}$ selects samples with lowest ratio.

\paragraph{Advanced Baselines.} For IFD, Superfiltering, SelectIT, and ZIP, we follow their original implementations and hyperparameters as described in their respective papers. All methods select 10\% of the training data (20\% for code domain).

\begin{figure*}[t]
    \centering
    \begin{minipage}[b]{0.25\textwidth}
        \centering
        \includegraphics[width=\textwidth]{figs/diff_entropy_general.pdf}
    \end{minipage}\hfill
    \begin{minipage}[b]{0.25\textwidth}
        \centering
        \includegraphics[width=\textwidth]{figs/diff_entropy_med.pdf}
    \end{minipage}\hfill
    \begin{minipage}[b]{0.25\textwidth}
        \centering
        \includegraphics[width=\textwidth]{figs/diff_entropy_math.pdf}
    \end{minipage}\hfill
    \begin{minipage}[b]{0.25\textwidth}
        \centering
        \includegraphics[width=\textwidth]{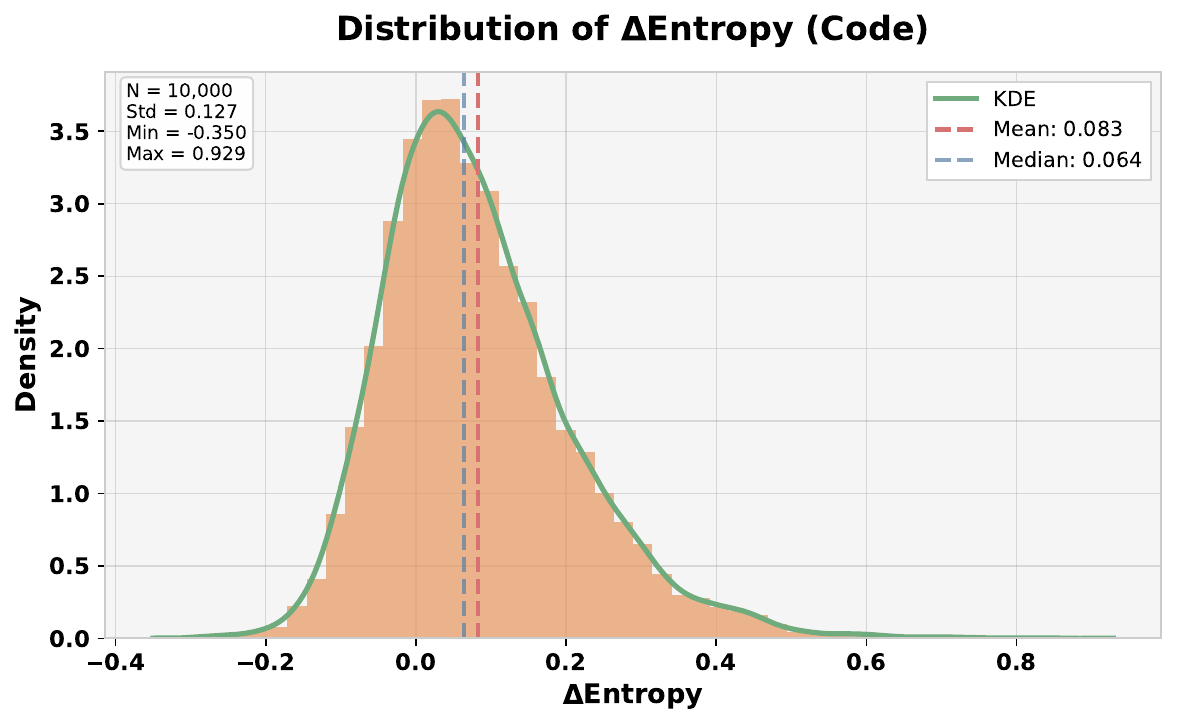}
    \end{minipage}
    \caption{\textbf{Domain-adaptive entropy dynamics reveal distinct learning patterns.} We measure entropy differences between base and calibrated instruction-tuned models across four domains. General instruction-following and medical QA exhibit entropy \textit{decrease} (\textbf{cognitive compression}), while mathematical reasoning and code generation exhibit entropy \textit{increase} (\textbf{cognitive expansion}). This domain-dependent pattern motivates our unified selection principle based on differential entropy.}
    \label{fig:domain_entropy_dynamics}
\end{figure*}

\section{Correlation Analysis Metrics}
\label{app:metrics}

We systematically evaluate how our core selection indicator, $\Delta H$, relates to other model and data properties. Figure~\ref{fig:correlation} summarizes all correlation results. All metrics considered are as follows:

\paragraph{Base Model Metrics} (computed using the initial model $\pi_{\text{base}}$): 
\begin{itemize}
    \item \textbf{Base NLL} ($\mathcal{L}_{\text{base}}$): average negative log-likelihood per token
    \item \textbf{Base Entropy} ($H_{\text{base}}$): average per-token output entropy (see Section~\ref{sec:state-diff})
\end{itemize}

\paragraph{Calibrated Model Metrics} (computed after warmup fine-tuning, using $\pi_{\text{inst}}$):
\begin{itemize}
    \item \textbf{Calib NLL} ($\mathcal{L}_{\text{inst}}$)
    \item \textbf{Calib Entropy} ($H_{\text{inst}}$)
\end{itemize}

\paragraph{Difference Metrics} (model-state gap, selection signal):
\begin{itemize}
    \item \textbf{Diff NLL} ($\Delta$NLL): change in NLL (Equation~\eqref{eq:delta_nll})
    \item \textbf{Diff Entropy} ($\Delta H$): entropy change (Equation~\eqref{eq:delta_h})
\end{itemize}

\paragraph{Auxiliary Statistics:}
\begin{itemize}
    \item \textbf{Instruction length} (Inst Len)
    \item \textbf{Response length} (Resp Len)
    \item \textbf{Length ratios}: Resp/Inst, Inst/Resp
\end{itemize}

\paragraph{Complexity Proxies:}
\begin{itemize}
    \item \textbf{Code:} XCoder~\cite{wang2024codellm} (code complexity/correctness)
    \item \textbf{Math:} pass@k rates (difficulty proxy)
    \item \textbf{General/Medical:} Instagger~\cite{lu2023instag} tag counts (instruction complexity)
\end{itemize}

Across all domains, we observe that $\Delta H$ is only weakly correlated with the above complexity proxies (see Figure~\ref{fig:correlation}). This supports that $\Delta H$ mainly captures model alignment and learnability rather than surface-level task difficulty.

\section{Training Hyperparameters}
\label{app:training_details}

\paragraph{Training and Evaluation.} 
All models use full-parameter supervised fine-tuning with AdamW optimizer, cosine learning rate decay, 
and 5\% warmup. For \mname, we set warmup ratio $\alpha = 0.1$, bi-directional reject ratio $\gamma = 
0.1$, and selection ratio $\beta = 0.1$ except for code, where $\beta = 0.2$. We evaluate mathematics 
on Math500~\citep{hendrycks2021measuring}, Minerva~\citep{lewkowycz2022solving}, OlympiadBench~\citep
{ai_mathematical_olympiad_2024}, AIME 2024~\citep{aime2024dataset}, and AMC 2023~\citep
{amc2023dataset} using accuracy; general instruction-following on Alpaca-Eval~\citep{alpaca_eval} 
using length-controlled win rate (LC Win) and raw win rate (WR); medical QA on MMLU-medical~\citep
{hendrycks2020measuring}, MedQA~\citep{jin2021disease}, and MedMCQA~\citep{pal2022medmcqa} using 
accuracy; and code generation on HumanEval/+~\citep{chen2021evaluatinglargelanguagemodels}, MBPP/
+~\citep{austin2021program}, and BigCodeBench~\citep{zhuo2024bigcodebench} using pass@k. Training 
hyperparameters are in Appendix~\ref{app:training_details}.

Table~\ref{tab:training_hyperparams} summarizes the training hyperparameters used across all domains.

\begin{table}[h]
\centering
\small

\begin{tabular}{lcccc}
\toprule
\textbf{Hyperparameter} & \textbf{Math} & \textbf{General} & \textbf{Medical} & \textbf{Code} \\
\midrule
Learning Rate & 5e-5 & 2e-5 & 2e-5 & 1e-5 \\
Batch Size & 256 & 64 & 64 & 256 \\
Epochs & 3 & 10 & 3 & 5 \\
Warmup Ratio & 0.05 & 0.05 & 0.05 & 0.05 \\
Weight Decay & 0.01 & 0.01 & 0.01 & 0.01 \\
Max Seq Length & 2048 & 2048 & 512 & 4096 \\
\bottomrule
\end{tabular}
\caption{Training hyperparameters for different domains.}
\label{tab:training_hyperparams}
\end{table}

All experiments use full-parameter fine-tuning (not LoRA) with AdamW optimizer, cosine learning rate decay, and gradient clipping at 1.0. Training is conducted on 8×A100 (80GB) GPUs with mixed precision (bf16).

\section{Algorithm Details}
\label{app:algorithm}

Algorithm~\ref{alg:instructdiff} presents the complete pseudo-code for \textsc{\mname}. The algorithm takes as input the full dataset $\mathcal{D}$, base model $\pi_{\text{base}}$, three hyperparameter ratios ($\alpha$ for warmup size, $\beta$ for final selection size, $\gamma$ for bi-directional filtering), and the number of iterations $K$. It returns the final selected subset $\mathcal{D}'^{(K)}$ of size $\beta \cdot N$.

The algorithm operates in two main stages. In Stage 1 (lines 1-2), we perform calibration by randomly sampling a warmup subset and fine-tuning the base model to obtain the initial measuring instrument $\pi_{\text{inst}}^{(0)}$. In Stage 2 (lines 3-11), we iterate $K$ times, each iteration computing distributional gaps for all samples (lines 4-6), applying bi-directional NLL filtering to retain the learnable middle range (line 7), selecting samples based on entropy difference (line 8), and optionally updating the calibration model for the next iteration (lines 9-11).

\begin{algorithm}[t]
\caption{\textsc{\mname}: Iterative Model-State Difference Selection}
\label{alg:instructdiff}
\begin{algorithmic}[1]
\REQUIRE Dataset $\mathcal{D}$, base model $\pi_{\text{base}}$, warmup ratio $\alpha$, selection ratio $\beta$, bi-reject ratio $\gamma$, iterations $K$
\ENSURE Selected subset $\mathcal{D}'$

\STATE // \textbf{Stage 1: Calibration}
\STATE $\mathcal{D}_{\text{warmup}} \leftarrow \text{RandomSample}(\mathcal{D}, \alpha \cdot N)$
\STATE $\pi_{\text{inst}}^{(0)} \leftarrow \text{FineTune}(\pi_{\text{base}}, \mathcal{D}_{\text{warmup}})$

\STATE // \textbf{Stage 2: Iterative Selection}
\FOR{$k = 1$ to $K$}
    \FOR{$(x_i, y_i) \in \mathcal{D}$}
        \STATE Compute $\Delta\text{NLL}_i \leftarrow \text{NLL}_{\text{inst}}^{(k-1)}(x_i, y_i) - \text{NLL}_{\text{base}}(x_i, y_i)$
        \STATE Compute $\Delta H_i \leftarrow H_{\text{base}}(x_i, y_i) - H_{\text{inst}}^{(k-1)}(x_i, y_i)$
    \ENDFOR
    
    \STATE // Bi-directional NLL filtering
    \STATE $\mathcal{D}_{\text{filtered}} \leftarrow \{(x_i, y_i) \in \mathcal{D} : q_{\gamma} \leq \Delta\text{NLL}_i \leq q_{1-\gamma}\}$
    
    \STATE // Entropy-based selection
    \STATE $\mathcal{D}'^{(k)} \leftarrow \text{SelectLowest}(\mathcal{D}_{\text{filtered}}, \beta \cdot N, \text{by } \Delta H)$
    
    \STATE // Update calibration model
    \IF{$k < K$}
        \STATE $\pi_{\text{inst}}^{(k)} \leftarrow \text{FineTune}(\pi_{\text{base}}, \mathcal{D}'^{(k)})$
    \ENDIF
\ENDFOR

\RETURN $\mathcal{D}'^{(K)}$
\end{algorithmic}
\end{algorithm}

\section{Case Studies}

We present representative selected and rejected samples in a unified visualization format to qualitatively illustrate our selection criteria.

\begin{figure*}[htbp]
    \centering
    \begin{tcolorbox}[title={Selected Sample: Low $\Delta H$, Moderate $\Delta$NLL}]
    \small
    \textbf{Instruction}\\
    List five potential stakeholders that could be involved in the design of a new regional park.\\[4pt]
    
    \textbf{Response}\\
    Some potential stakeholders that could be involved in the design of a new regional park are:\\
    1. Local government agencies: These agencies are responsible for managing and developing public spaces, including parks, within their jurisdiction.\\
    2. Community groups: Representatives from local community groups that can provide insight into the needs and concerns of the residents who will use the park.\\
    3. Environmental organizations: Organizations that focus on environmental conservation and sustainability can provide valuable input on how to design the park in a way that benefits both the community and the environment.\\
    4. Local business owners: Local businesses near the park could benefit from increased foot traffic and may have a vested interest in the park's design and success.\\
    5. Landscape architects / urban planners: Professionals in these fields have expertise in designing functional and aesthetically pleasing public spaces.\\[4pt]
    
    \textbf{Metrics}\\
    $\Delta$NLL = 0.06 \quad
    $\Delta$ H = 0.16 \quad
    $\Delta$PPL = 0.15\\[4pt]
    
    \textbf{Interpretation}\\
    This example exhibits low entropy (high model confidence) and a moderate NLL gap, indicating a stable yet informative learning signal. It lies within the desired \emph{activation zone} and is therefore retained.
    \end{tcolorbox}
    \caption{Example of a selected training sample with balanced uncertainty and learning signal}
    \label{fig:case_selected}
\end{figure*}

\begin{figure*}[htbp]
    \centering
    \begin{tcolorbox}[title=Rejected Sample: Excessively High $\Delta$NLL]
    \small
    \textbf{Instruction}\\
    Create an antonym for the word ``Early''. Output the antonym.\\[4pt]
    
    \textbf{Response}\\
    Late.\\[4pt]
    
    \textbf{Metrics}\\
    $\Delta$NLL = 4.14 \quad
    $\Delta$ H = 2.19 \quad
    $\Delta$PPL = 201.24\\[4pt]
    
    \textbf{Interpretation}\\
    Despite being factually correct, this sample induces an abnormally large NLL and perplexity gap, suggesting unstable gradients and potential over-penalization during training. Such samples are filtered to prevent noisy updates.
    \end{tcolorbox}
    \caption{Rejected example due to excessively high NLL gap and unstable learning signal}
    \label{fig:case_reject_high_nll}
\end{figure*}

\begin{figure*}[htbp]
    \centering
    \begin{tcolorbox}[title=Rejected Sample: Degenerate Low $\Delta$NLL]
    \small
    \textbf{Instruction}\\
    Output the longest palindrome in a sentence. This is sentence with a palindrome.\\[4pt]
    
    \textbf{Response}\\
    a palindrome\\[4pt]
    
    \textbf{Metrics}\\
    $\Delta$NLL = -2.84 \quad
    $\Delta$ H = 1.79 \quad
    $\Delta$PPL = -552.69\\[4pt]
    
    \textbf{Interpretation}\\
    This sample yields an anomalously low (negative) NLL and perplexity gap, indicating trivial memorization or metric artifacts rather than meaningful learning. Such samples contribute little to model improvement and are therefore discarded.
    \end{tcolorbox}
    \caption{Rejected example with degenerate low NLL gap and limited training value}
    \label{fig:case_reject_low_nll}
\end{figure*}

This sample has extremely high gaps, indicating it is beyond the model's current capability. Training on such samples introduces harmful noise.

\end{document}